\documentclass[letterpaper, 10pt, conference]{ieeeconf}
\IEEEoverridecommandlockouts
\usepackage{cite}
\usepackage{amsmath,amssymb,amsfonts}
\usepackage{algorithmic}
\usepackage{algorithm}
\usepackage{graphicx}
\usepackage{textcomp}
\usepackage{xcolor}

\title{\LARGE \bf
Beyond Imitation: Reinforcement Learning Fine-Tuning for Adaptive Diffusion Navigation Policies
}

\author{Junhe Sheng$^{1}$, Ruofei Bai$^{1,2}$, Kuan Xu$^{1}$, Ruimeng Liu$^{1}$, Jie Chen$^{3}$, Shenghai Yuan$^{1}$, Wei-Yun Yau$^{2}$, Lihua Xie$^{1}$%
  \thanks{$^{1}$Nanyang Technological University, Singapore.}%
  \thanks{$^{2}$A*STAR, Singapore.}%
  \thanks{$^{3}$National University of Singapore, Singapore.}%
}
\begin{document}
\maketitle
\thispagestyle{empty}
\pagestyle{empty}

\begin{abstract}
Diffusion-based robot navigation policies trained on large-scale imitation learning datasets, can generate multi-modal trajectories directly from the robot's visual observations, bypassing the traditional localization–mapping–planning pipeline and achieving strong zero-shot generalization. However, their performance remains constrained by the coverage of offline datasets, and when deployed in unseen settings, distribution shift often leads to accumulated trajectory errors and safety-critical failures. Adapting diffusion policies with reinforcement learning is challenging because their iterative denoising structure hinders effective gradient backpropagation, while also making the training of an additional value network computationally expensive and less stable.
To address these issues, we propose a reinforcement learning fine-tuning framework tailored for diffusion-based navigation. The method leverages the inherent multi-trajectory sampling mechanism of diffusion models and adopts Group Relative Policy Optimization (GRPO), which estimates relative advantages across sampled trajectories without requiring a separate value network. 
To preserve pretrained representations while enabling adaptation, we freeze the visual encoder and selectively update the higher decoder layers and action head, enhancing safety-aware behaviors through online environmental feedback. On the PointGoal task in Isaac Sim, our approach improves the Success Rate from 52.0\% to 58.7\% and SPL from 0.49 to 0.54 on unseen scenes, while reducing collision frequency. Additional experiments show that the fine-tuned policy transfers zero-shot to a real quadruped platform and maintains stable performance in geometrically out-of-distribution environments, suggesting improved adaptability and safe generalization to new domains.
\end{abstract}


\section{Introduction}

Autonomous robot navigation in complex, unmapped environments remains a fundamental challenge in embodied AI. Recently, large-scale data-driven foundation models \cite{shah2023vint, shah2023gnm, cai2025navdp} have achieved remarkable zero-shot generalization by framing navigation as a behavior cloning problem over massive offline datasets. Despite their success, these pretrained models are inherently bounded by the coverage of their training data. When deployed in novel environments with unseen obstacle layouts, tight corridors, or different robot dynamics, they frequently encounter distribution shifts that result in safety-critical failures: the robot misjudges clearances, plans paths that graze or collide with obstacles, and produces unreliable trajectories that oscillate or diverge near tight spaces. These failures expose a fundamental gap between learned policies and safe real-world deployment---ensuring that the outputs of learned navigation models are safe and reliable remains an open and pressing challenge for the community. Online adaptation is therefore essential, not merely to improve task success, but to close this safety gap.

Reinforcement Learning (RL) provides a natural mechanism for adapting these pretrained policies through online environmental interaction. However, applying existing RL methods to diffusion-based foundation models is highly non-trivial. Recent works attempting to fine-tune diffusion policies with RL typically rely on standard actor-critic architectures (e.g., DPPO \cite{ren2025diffusion}). Because a diffusion policy generates actions through an iterative denoising process, the actor must be split into a frozen pretrained copy and a trainable fine-tuned copy, and training an additional critic network alongside both copies significantly increases memory consumption. Furthermore, the critic must be trained from scratch without any pretrained initialization, making value estimation noisy and slow to converge, which can destabilize the overall policy optimization.

To overcome these bottlenecks, we propose a novel, critic-free RL fine-tuning framework specifically designed to adapt pretrained diffusion navigation models without requiring additional expert demonstrations. Rather than evaluating individual actions with a separate value network, we formulate the diffusion policy as a trajectory-level distribution $\pi_\theta(\tau \mid o_t)$. Recognizing that diffusion sampling naturally produces multiple candidate trajectories per rollout, we leverage Group Relative Policy Optimization (GRPO) \cite{shao2024deepseekmath} to compute relative advantages directly within each candidate set. This trajectory-level optimization completely eliminates the need for a separate critic network, drastically reducing memory overhead and stabilizing the training process. Crucially, by incorporating collision penalties and clearance constraints through online environmental interaction, the framework enables the policy to acquire safety-aware planning behaviors---such as maintaining adequate margins around obstacles and avoiding near-miss paths---that the original offline policy often fails to maintain under distribution shift.

To prevent catastrophic forgetting of the pretrained visual and geometric priors, we further introduce a selective fine-tuning strategy that freezes the perception backbone and the lower Transformer decoder layers, updating only the upper layers and action head. This combination of critic-free optimization and selective freezing boosts the success rate from 52.0\% to 58.7\% in unseen simulation environments, producing safer and more reliable trajectories, while preserving zero-shot transferability to a physical quadruped robot—all without any real-world training.

Our contributions are summarized as follows:
\begin{itemize}
    \item We formulate diffusion navigation policies as sampling from trajectory distributions and derive policy gradients through the denoising chain.
    \item We adapt GRPO to exploit the multi-sample nature of diffusion models for variance-reduced, critic-free optimization.
    \item We show that selective parameter freezing during RL fine-tuning is necessary to preserve the geometric priors from pretraining.
    \item We present the first zero-shot cross-simulator and cross-embodiment transfer study of an RL-adapted diffusion navigation model.
\end{itemize}

\section{Related Work}
Our work is situated at the intersection of two major research areas: large-scale, data-driven navigation, and the adaptation of generative models using reinforcement learning.
\subsection{Data-Driven Navigation and Foundation Models}
The paradigm for embodied navigation has increasingly shifted towards large-scale learning \cite{wu2024embodied, liu2025embodied}. Early end-to-end RL approaches such as DD-PPO \cite{wijmans2020ddppo} demonstrated that agents could learn navigation from scratch, but required billions of interactions and struggled to generalize across distinct domains. Foundation models based on large-scale imitation learning---such as GNM \cite{shah2023gnm}, ViNT \cite{shah2023vint}, and vision-language-action architectures \cite{zitkovich2023rt2, kim2025openvla}---address this by learning reusable navigation priors from diverse offline datasets, yielding strong zero-shot performance.
Diffusion-based policies \cite{chi2025diffusion, yu2024trajectory, ze20243d} further advance this paradigm by modeling multimodal trajectory distributions via conditional denoising, enabling more flexible and robust plans. NavDP \cite{cai2025navdp}, which serves as our base model, exemplifies this success with impressive sim-to-real transfer. Despite these strengths, all such models are fundamentally constrained by their offline training data and lack mechanisms to adapt or recover from errors in out-of-distribution scenarios. Our work directly addresses this gap by enabling online adaptation.
\subsection{Reinforcement Learning for Generative and Diffusion Policies}
The challenge of adapting pretrained generative policies has recently drawn significant attention. Several works have explored combining RL with diffusion models, but they differ critically from our approach. Methods like DPPO \cite{ren2025diffusion} and QVPO \cite{ding2024diffusion} apply actor-critic RL frameworks to diffusion policies. However, they require training an independent value network alongside the diffusion actor. Since the critic lacks pretrained initialization and must be learned from scratch, its value estimates can be noisy during early training, potentially destabilizing the policy optimization. Moreover, standard DPPO maintains both a frozen pretrained actor and a trainable fine-tuned copy for selective denoising, further increasing memory overhead compared to critic-free alternatives.
Other works have explored offline RL or trajectory optimization techniques for diffusion models \cite{stamatopoulou2024dippest, qin2025integrating}. These methods can be effective but typically rely on the assumption that either a static dataset of expert-level interactions or an accurate dynamics model is available. This assumption is often violated in real-world deployment scenarios, where the goal is to continuously learn and adapt in a new environment without access to additional expert data.
In contrast, our work operates at the trajectory level and employs GRPO \cite{shao2024deepseekmath} to compute relative advantages directly from diffusion-sampled candidates, completely eliminating the value network and enabling stable, memory-efficient online adaptation without pre-existing expert data.

\section{Preliminaries}

\subsection{Problem Formulation}
We formulate autonomous navigation as a goal-conditioned decision process. At each timestep $t$, the agent receives an observation $o_t$ comprising RGB-D images and a relative goal vector $g_t \in \mathbb{R}^2$. The objective is to reach the goal while avoiding obstacles. Following recent foundation models \cite{cai2025navdp}, we adopt a receding-horizon control strategy: the policy $\pi_\theta( \tau_t \mid o_t )$ predicts a future trajectory $\tau_t = \{x_1, \dots, x_H\} \in \mathbb{R}^{H \times 2}$ of $H$ waypoints in the robot's local frame. A low-level controller then tracks the first few waypoints to generate velocity commands.

\subsection{Diffusion-Based Trajectory Generation}
We model the policy as a conditional diffusion process \cite{chi2025diffusion}. The trajectory distribution $\pi_\theta(\tau_t \mid o_t)$ is represented by a reverse denoising chain. Starting from Gaussian noise $\tau^K \sim \mathcal{N}(\mathbf{0}, \mathbf{I})$, the model iteratively refines the trajectory over $K$ steps:
\begin{equation}
    \tau^{k-1} = \mu_\theta(\tau^k, o_t, k) + \sigma_k \mathbf{z}, \quad \mathbf{z} \sim \mathcal{N}(\mathbf{0}, \mathbf{I}),
\end{equation}
where $k$ denotes the diffusion step (distinct from time $t$). The mean $\mu_\theta$ is parameterized by a noise prediction network $\epsilon_\theta(\tau^k, k, o_t)$, typically trained via a simple mean-squared error loss on the noise term:
\begin{equation}
    \mathcal{L}_{\text{diff}} = \mathbb{E}_{k, \tau^0, \epsilon} \left[ \| \epsilon - \epsilon_\theta(\tau^k, k, o_t) \|^2 \right],
\end{equation}
where $\epsilon \sim \mathcal{N}(\mathbf{0}, \mathbf{I})$ is the Gaussian noise sampled at each step.

\subsection{Policy Optimization via Trajectory Likelihood}
To apply RL to this generative policy, we need the likelihood of a generated trajectory. The log-probability of a sampled trajectory $\tau^0$ decomposes as the sum over all reverse-step transitions:
\begin{equation}
    \log \pi_\theta(\tau^0 \mid o_t) = \sum_{k=1}^{K} \log p_\theta(\tau^{k-1} \mid \tau^k, o_t).
\end{equation}
Since each transition $p_\theta(\cdot \mid \cdot)$ is Gaussian, $\nabla_\theta \log \pi_\theta$ can be computed by backpropagating through $\epsilon_\theta$ at each step $k$. This allows us to optimize the expected trajectory return $J(\theta) = \mathbb{E}_{\tau \sim \pi_\theta}[R(\tau)]$ using standard policy gradient methods.

\subsection{Pretrained Foundation Model}
We initialize our policy with Navigation Diffusion Policy (NavDP) \cite{cai2025navdp}, a diffusion-based navigation model pretrained on large-scale simulation data. NavDP uses a vision encoder for RGB-D inputs and a Diffusion Transformer (DiT) decoder to generate trajectories. NavDP generalizes well zero-shot, but its performance is bounded by the offline training distribution. Our goal is to adapt $\theta$ online to improve task success in novel environments.

\section{Method}

\subsection{Reinforcement Learning Fine-Tuning Framework}

\begin{figure}[t]
    \centering
    \includegraphics[width=0.95\linewidth]{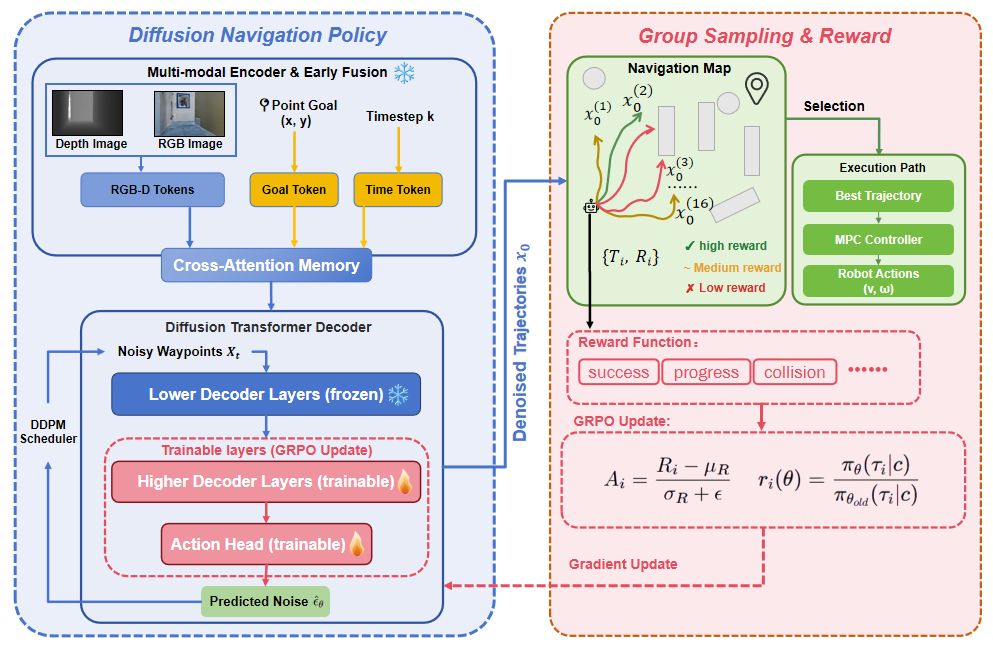}
    \caption{Overview of our framework.}
    \label{fig:framework}  
\end{figure}

We adapt a pretrained diffusion-based navigation policy through RL, without requiring additional expert demonstrations. An overview of our framework is shown in Fig.~\ref{fig:framework}.

At each timestep, given observation condition $o_t$, the policy samples a group of $G=16$ trajectory candidates $\{\tau_i\}_{i=1}^{G} \sim \pi_\theta(\cdot \mid o_t)$. Each trajectory is evaluated using a reward function $R_i = R(\tau_i)$, and the policy is optimized to maximize expected reward:

\begin{equation}
\max_\theta
\mathbb{E}_{\tau \sim \pi_\theta}
[R(\tau)].
\end{equation}

\subsection{Reward Function}

We design a trajectory-level reward that encourages goal reaching, efficient progress, collision avoidance, and smooth motion. Importantly, the $G$ trajectory candidates are evaluated analytically in memory using the current observation, without executing them in the environment. We use the current depth image and camera intrinsics to build a local 2D occupancy grid $\mathcal{O}_{occ}$ and its inflated version $\mathcal{O}_{infl}$. Each trajectory is projected onto this grid to compute collision penalties.
Specifically, let $\{p_h\}_{h=1}^H$ denote the predicted waypoints and $g$ the target goal position. The total reward is a weighted sum:

\begin{equation}
R(\tau) = \sum_{i} w_i R_i.
\end{equation}

The individual components $R_i$ and their respective weights $w_i$ are summarized in Table~\ref{tab:reward}.

\begin{table}[h]
\centering
\caption{Reward function components.}
\label{tab:reward}
\begin{tabular}{ccc}
\hline
\textbf{Component} & \textbf{Definition} & \textbf{Weight} \\
\hline
Success & $\mathbb{I}(\|p_H - g\|_2 < 0.3)$ & +10.0 \\
Progress & $\max(0, \|p_0 - g\|_2 - \|p_H - g\|_2)$ & +3.0 \\
Collision & $\frac{1}{H}\sum_{h=1}^H \mathbb{I}(p_h \in \mathcal{O}_{occ})$ & -5.0 \\
Inflated Coll. & $\frac{1}{H}\sum_{h=1}^H \mathbb{I}(p_h \in \mathcal{O}_{infl})$ & -1.0 \\
Distance & $\|p_H - g\|_2$ & -0.1 \\
Smoothness & $\frac{1}{H{-}1}\sum_{h=1}^{H-1} \|p_{h+1} - p_h\|_2$ & -0.1 \\
Zig-zag & $\frac{1}{H{-}1}\sum_{h=1}^{H-2} \mathbb{I}(\text{sgn}(\Delta y_{h+1}) \neq \text{sgn}(\Delta y_h))$ & -0.05 \\
\hline
\end{tabular}
\end{table}

\subsection{Group Relative Policy Optimization}

To enable efficient optimization of diffusion policies, we adopt Group Relative Policy Optimization (GRPO) \cite{shao2024deepseekmath}, which estimates relative advantages within trajectory groups.
Given rewards $\{R_i\}_{i=1}^{G}$ sampled under the same observation condition, the advantage is computed using group normalization:

\begin{equation}
A_i =
\frac{
R_i - \mu_R
}{
\sigma_R + \epsilon
},
\end{equation}

\noindent where $\mu_R$ and $\sigma_R$ denote the mean and standard deviation of rewards within the group.
The policy update is performed using a PPO-style clipped surrogate objective \cite{schulman2017proximal}:

\begin{equation}
r_i(\theta)
=
\frac{
\pi_\theta(\tau_i \mid o_t)
}{
\pi_{\theta_{old}}(\tau_i \mid o_t)
},
\end{equation}

\begin{equation}
\mathcal{L}_{RL}
=
-
\mathbb{E}
\left[
\min(
r_i A_i,
\text{clip}(r_i, 1-\epsilon_{clip}, 1+\epsilon_{clip}) A_i
)
\right].
\end{equation}

This objective encourages trajectories with higher relative rewards while preventing excessively large policy updates, where $\epsilon_{clip}$ is a clipping hyperparameter.

\textbf{Importance Ratio via Denoising Chain.}
Computing the importance ratio $r_i(\theta)$ for diffusion policies requires some care. Since the trajectory likelihood is a product of all reverse-step transition probabilities (Eq.~3), the log-ratio is obtained by summing over the full $K$-step denoising chain:

\begin{equation}
\begin{split}
\log r_i(\theta)
= \sum_{k=1}^{K}
\Big[
&\log p_\theta(\tau^{(k-1)}_i \mid \tau^{(k)}_i, o_t) \\
&- \log p_{\theta_{old}}(\tau^{(k-1)}_i \mid \tau^{(k)}_i, o_t)
\Big],
\end{split}
\end{equation}

\noindent where each $\log p_\theta$ is evaluated as a Gaussian log-likelihood with mean given by the DDPM posterior (Eq.~1). During training, we cache the per-step noise predictions from the sampling policy $\pi_{\theta_{old}}$ to enable efficient computation of log-ratios without additional forward passes through the old model.

\subsection{KL-Free Optimization}

A common practice in RL fine-tuning is to add a KL divergence penalty $\beta D_{KL}(\pi_\theta || \pi_{ref})$ to keep the policy close to the pretrained reference. We set $\beta = 0$ and rely on the clipped surrogate objective alone for stability. The clipping already bounds the policy ratio at each step, and the selective parameter freezing (Section~IV-F) limits how much the model can change. Our ablation (Section~V-D) confirms that removing the KL penalty improves performance, likely because it allows the policy to adjust its behavior for obstacle configurations underrepresented in the pretraining data.

\subsection{Selective Fine-Tuning of Foundation Model}

The perception encoder extracts visual features from RGB-D observations, and the decoder generates trajectories conditioned on these features. Updating all parameters during RL risks catastrophic forgetting of the visual representations from pretraining. To avoid this, we partition the model parameters:

\begin{equation}
\theta =
\{\theta_{enc}, \theta_{dec}^{frozen}, \theta_{dec}^{train}, \theta_{head}\},
\end{equation}

\noindent where $\theta_{enc}$ is the perception encoder, $\theta_{dec}^{frozen}$ is the first $L-N$ decoder layers, $\theta_{dec}^{train}$ is the last $N$ decoder layers, and $\theta_{head}$ is the action head. We freeze $\theta_{enc}$ and $\theta_{dec}^{frozen}$, and only update $\theta_{dec}^{train}$ and $\theta_{head}$.

\textbf{Choice of decoder layers to update.}
This design is motivated by the hierarchical nature of Transformer decoders: lower layers encode universal spatio-temporal primitives shared across environments, while upper layers capture task-specific trajectory generation behaviors that are more amenable to adaptation. Freezing the lower layers is therefore preferable to preserve the core geometric representations built during large-scale pretraining. Updating only a small number of top layers strikes a balance between plasticity and stability; the specific value of $N$ is validated empirically in Section~V-D.

\textbf{Choice of denoising steps for gradient computation.}
Not all denoising steps contribute equally to the final trajectory. Early steps, which transform pure Gaussian noise into coarse structure, are dominated by high-variance noise and provide little geometric information. In contrast, the final steps refine the trajectory details and are most predictive of the output. Rather than summing over the full $K$-step chain when computing the importance ratio (Eq.~9), we therefore restrict the gradient signal to the last $K$ steps of the chain. The optimal value of $K$ is evaluated in Section~V-D.

During training, gradient updates are applied exclusively to the trainable parameters $\{\theta_{dec}^{train}, \theta_{head}\}$, with all other parameters frozen.
\subsection{Training Procedure}

We adopt an offline, buffered GRPO training pipeline that alternates between data collection and policy optimization across multiple iterations. The complete procedure is described in Algorithm~\ref{alg:training}.

\begin{algorithm}[t]
\caption{Offline Buffered GRPO Fine-Tuning}
\label{alg:training}
\begin{algorithmic}[1]
\REQUIRE Pretrained policy $\pi_0$, scene pool $\mathcal{S}$, total iterations $M$
\STATE Initialize policy $\pi \leftarrow \pi_0$
\FOR{iteration $m = 1, 2, ..., M$}
    \STATE Select scene subset $\mathcal{S}_m \subset \mathcal{S}$ (rolling window)
    \STATE Initialize disk-based replay buffer $\mathcal{B} \leftarrow \emptyset$
    \FOR{each scene $s \in \mathcal{S}_m$}
        \STATE Collect $N_{ep}$ episodes using $\pi$ in scene $s$
        \STATE Compute trajectory rewards $R(\tau)$ and cache denoising-chain log-probabilities
        \STATE Store experience tuples in $\mathcal{B}$
    \ENDFOR
    \FOR{epoch $= 1, 2, ..., E$}
        \FOR{each mini-batch $\{(\tau_i, R_i, o_{t,i})\}$ from $\mathcal{B}$}
            \STATE Compute group advantages $A_i$ via Eq.~6
            \STATE Compute importance ratios via denoising chain (Eq.~9)
            \STATE Update $\{\theta_{dec}^{train}, \theta_{head}\}$ using $\mathcal{L}_{RL}$ (Eq.~8)
        \ENDFOR
    \ENDFOR
    \STATE Select best checkpoint based on windowed average reward
    \STATE Update $\pi \leftarrow$ best checkpoint, clear $\mathcal{B}$
\ENDFOR
\end{algorithmic}
\end{algorithm}

\textbf{Key implementation details.} In each iteration, the policy collects $N_{ep}=130$ episodes across 4 training scenes, stored in a disk-based replay buffer (capacity 128). Policy updates run for $E=2$ epochs with mini-batch size 64. The best checkpoint is selected via windowed average reward (window size 5), and the buffer is cleared before the next iteration. This offline buffered design decouples data collection from optimization, enabling stable training with a fixed $\pi_{\theta_{old}}$ per iteration.

\section{Experiments}

We evaluate our framework by addressing three core questions: (1) Does trajectory-level RL fine-tuning improve navigation performance over the pretrained foundation model? (2) Does selective fine-tuning preserve generalizable representations while mitigating catastrophic forgetting? (3) How do the specific components of our critic-free GRPO objective impact adaptation?

\subsection{Experimental Setup}

\subsubsection{Simulation Platform and Datasets}
\begin{figure}[h]
    \centering
    \includegraphics[width=\linewidth]{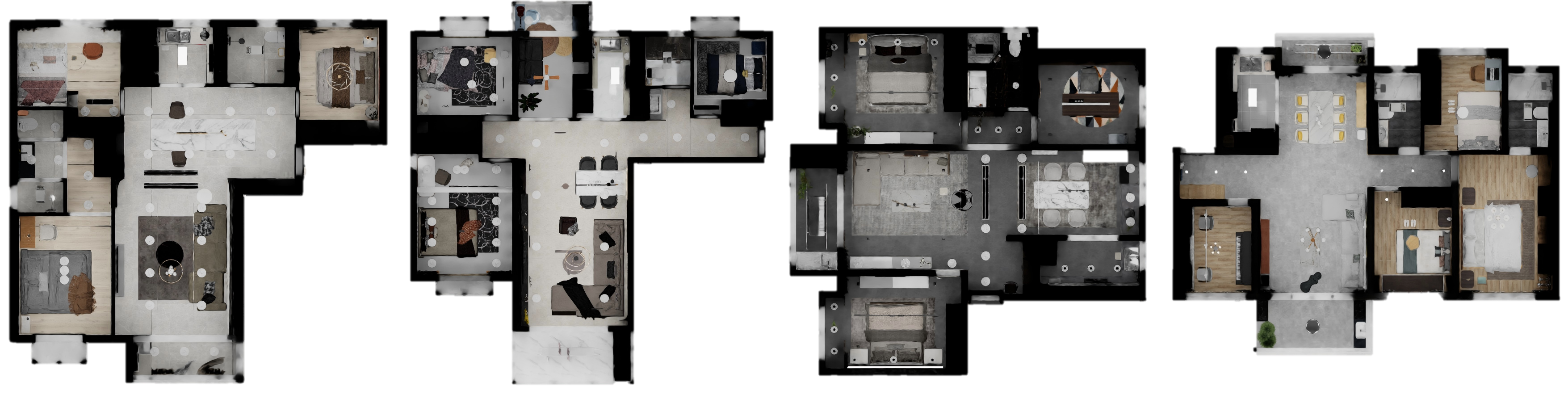}
    \caption{Representative high-fidelity residential environments from the InternScenes-Home dataset used for training and evaluation. Image adapted from \cite{cai2025navdp}.}
    \label{fig:home_scenes}
\end{figure}

All simulation experiments are conducted on a high-fidelity physics platform powered by Isaac Sim (4.2.0) and Isaac Lab (1.2.0)~\cite{mittal2025isaac}. To thoroughly assess navigation performance in cluttered, realistic indoor spaces, we utilize the high-quality scene assets from the InternScene-N1 dataset \cite{scene_n1}. Specifically, we deploy the robot in the \emph{InternScenes-Home} subset, which features diverse residential layouts with complex geometry and rich semantics. Representative frames from these Home environments are shown in Fig.~\ref{fig:home_scenes}. 

The simulation includes a Dingo wheeled mobile robot equipped with a front-facing RGB-D camera and contact sensors. At each timestep (10\,Hz control loop), the policy observes a temporal history of $K=8$ RGB-D frames to predict a trajectory of $H=24$ future waypoints, which are subsequently executed by a low-level MPC controller.

\subsubsection{Model Architecture and Selective Fine-Tuning Details}
We initialize our model using the pretrained Navigation Diffusion Policy (NavDP) \cite{cai2025navdp}. The full model has a total capacity of \textbf{173.59\,M parameters}, comprising two complete DepthAnythingV2 (vits) encoders~\cite{yang2024depth}, a 2-layer early fusion network, a 16-layer Diffusion Transformer (DiT) decoder, and a prediction head. 

Following our selective fine-tuning strategy (Section IV-E), we freeze the perception encoders and the first 11 DiT decoder layers. We only optimize the trainable parameters in the final 5 decoder layers (layers 11--15, which include the Self-Attention, Cross-Attention, and Feed-Forward Network blocks) and the action head. This strict layer-freezing reduces the trainable parameter count to just \textbf{11.83\,M parameters} (approximately 6.8\% of the total model). This lightweight configuration not only reduces VRAM requirements (peaking strictly under 17\,GB) but also effectively mitigates catastrophic forgetting of the pretrained visual features. 

\subsubsection{Training and Evaluation Protocol}
Training follows an iterative, scene-rolling procedure to prevent overfitting. In each of the 8 training iterations, the policy collects 130 episodes across a rolling window of 4 residential scenes (sampled from Home 11--19). The data is stored in a disk-based replay buffer (capacity of 128 episodes) to decouple data collection from optimization. GRPO updates are performed offline for 2 epochs with a batch size of 64 and a learning rate of $1 \times 10^{-5}$ ($\beta=0.0$). 

For evaluation, we report two standard point-goal navigation metrics: \textbf{Success Rate (SR)} measures the percentage of episodes in which 
the robot reaches within $1.5$\,m of the goal. \textbf{Success weighted by Path Length (SPL)} \cite{anderson2018evaluation} is defined as $S \cdot \frac{L}{\max(P,\, L)}$, where $S \in \{0,1\}$ is the success indicator, $L$ is the shortest-path length, and $P$ is the actual path traversed.

Crucially, we abandon randomized spawning to ensure strict fairness and reproducibility. Instead, all evaluation episodes use fixed start and goal poses defined explicitly within pre-generated USD scene files. We evaluate the models on 11 held-out, unseen scenes (Home 0--10) from the InternScenes-Home dataset, executing 101 distinct tasks per scene (1,111 total episodes). All evaluations are averaged over 3 global random seeds (1234, 42, 10). We compare against: \textbf{DD-PPO} \cite{wijmans2020ddppo}, an end-to-end RL agent trained from scratch; \textbf{iPlanner} \cite{yang2023iplanner}, a depth-based local planner trained via imitation learning; \textbf{ViPlanner} \cite{roth2024viplanner}, a semantic local planner using depth and semantic features; and \textbf{DPPO} \cite{ren2025diffusion}, actor-critic RL fine-tuning of a diffusion policy with an auxiliary value network.

\subsection{Main Results and Generalization Analysis}

Table~\ref{tab:main_results} summarizes the quantitative navigation performance. 
The environment is highly challenging, as evidenced by the end-to-end RL baseline (DD-PPO) achieving only a 12.1\% success rate in unseen scenes. While the zero-shot pretrained NavDP foundation model achieves a strong baseline of 52.0\%, it exhibits characteristic failure modes that reveal a fundamental safety limitation. Although NavDP incorporates safety awareness through offline contrastive learning with privileged information, its critic relies heavily on the training distribution. When deployed in unseen settings, this offline safety mechanism often fails to generalize: the robot underestimates its physical footprint, plans paths that appear clear but leave inadequate clearance, and scrapes past furniture edges or becomes wedged in narrow gaps. Since privileged safety annotations are unavailable at inference time and training-scene boundaries differ significantly from novel environments, the critic cannot reliably distinguish near-miss trajectories from safe ones---policy outputs can appear reasonable yet remain unsafe. Our RL fine-tuning addresses this gap by exposing the model to true safety boundaries through online interaction, enabling it to learn robust constraints---maintaining sufficient clearance and rejecting near-collision paths---that the offline critic cannot enforce in novel environments.

\begin{table}[h]
\centering
\caption{Quantitative results on Seen (Training) and Unseen (Test) environments. Metrics are reported as the mean over 3 seeds.}
\resizebox{\linewidth}{!}{
\begin{tabular}{cccc}
\hline
\textbf{Environment} & \textbf{Method} & \textbf{Success Rate} $\uparrow$ & \textbf{SPL} $\uparrow$ \\
\hline
\textbf{Seen} & NavDP & $0.601$ & $0.577$ \\
\textbf{Seen} & \textbf{Ours} & $\mathbf{0.687}$ & $\mathbf{0.628}$ \\
\hline
\textbf{Unseen} & DD-PPO \cite{wijmans2020ddppo} & $0.121$ & $0.362$ \\
\textbf{Unseen} & iPlanner \cite{yang2023iplanner} & $0.362$ & $0.345$ \\
\textbf{Unseen} & ViPlanner \cite{roth2024viplanner} & $0.402$ & $0.390$ \\
\textbf{Unseen} & NavDP & $0.520$ & $0.490$ \\
\textbf{Unseen} & DPPO \cite{ren2025diffusion} & $0.527$ & $0.488$ \\
\textbf{Unseen} & \textbf{Ours} & $\mathbf{0.587}$ & $\mathbf{0.537}$ \\
\hline
\end{tabular}
}
\label{tab:main_results}
\end{table}

Our RL fine-tuned framework (\textbf{Ours}) significantly outperforms the pretrained NavDP and all other baselines. Notably, while applying standard actor-critic RL to the diffusion model (DPPO \cite{ren2025diffusion}) improves over the zero-shot NavDP baseline (raising the success rate to 52.7\%), its performance still lags behind our critic-free GRPO approach (58.7\%). This performance gap suggests that maintaining an independent value network adds optimization complexity without matching the effectiveness of directly comparing trajectory returns within a group.

\textbf{Generalization Gap Analysis:} In seen environments, our RL fine-tuning improves the success rate by an absolute 8.6\% ($0.601 \to 0.687$) over NavDP. Crucially, this improvement transfers robustly to unseen environments, showing a 6.7\% absolute increase ($0.520 \to 0.587$). The narrow margin between the training and testing gains validates our hypothesis: by strictly updating only the upper decoder layers, the model learns generalizable collision-avoidance behaviors rather than merely memorizing the training layouts.

\textbf{Computational Cost:} Our fine-tuning process is efficient, requiring approximately 1 hour for pure training and 2 hours for data collection per iteration on a single NVIDIA RTX 6000 Ada GPU (total $\sim$3 hours per round). Inference latency remains consistent with the baseline ($\sim$500 ms/step), as the model architecture is unchanged.

\begin{figure}[h]
\centering
\includegraphics[width=\linewidth]{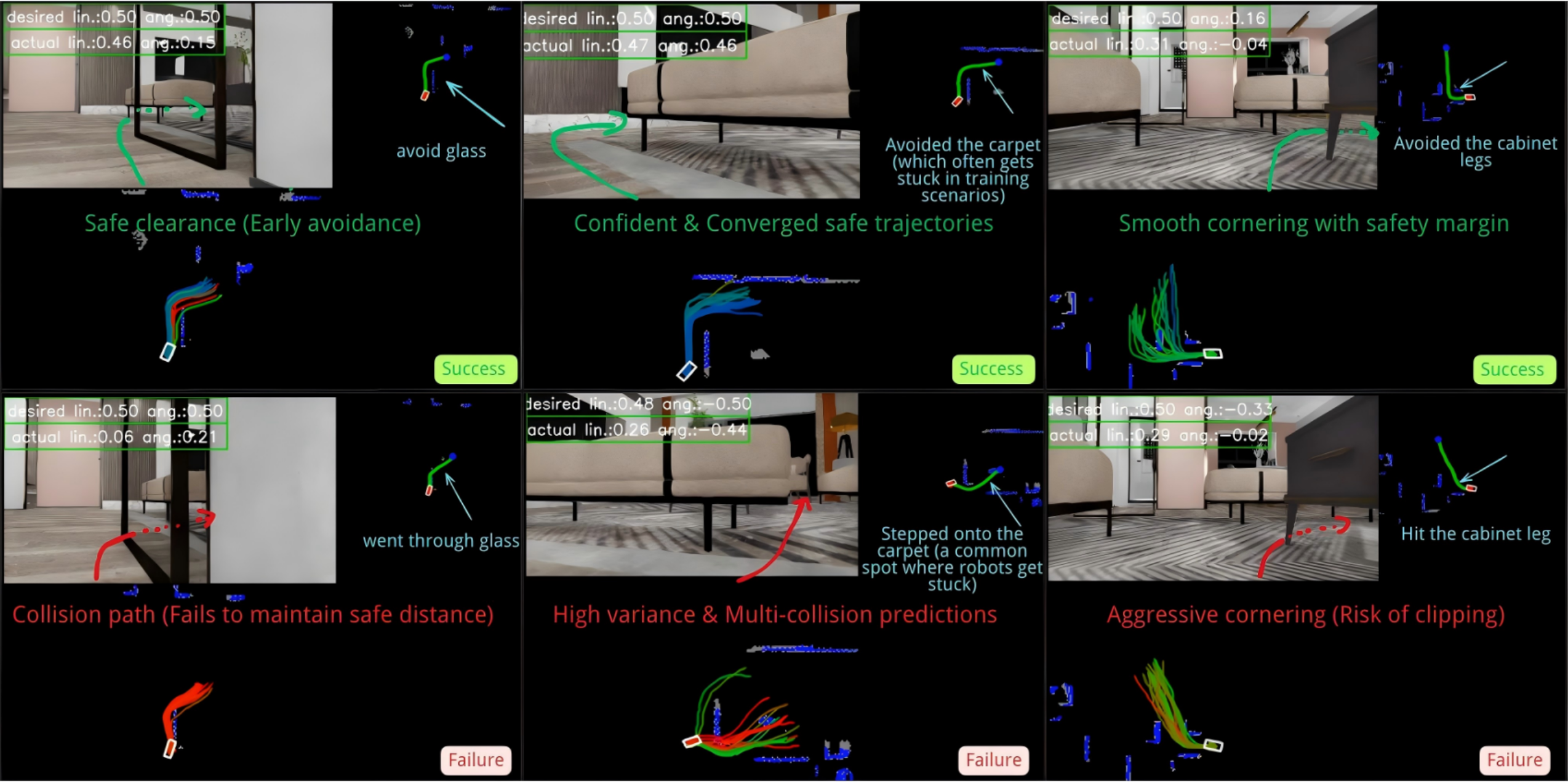}
\caption{Trajectory comparison. Our fine-tuned policy (top row, green) generates tightly converged, safe avoidance bundles. The NavDP baseline (bottom row, red) produces highly divergent and unsafe paths.}
\label{fig:trajectory}
\end{figure}

\textbf{Qualitative Analysis.}
Fig.~\ref{fig:trajectory} provides a detailed qualitative comparison of trajectory candidates. Our fine-tuned policy (top row, green) preemptively anticipates obstacles, producing tightly converged trajectory bundles with minimal variance. It successfully executes smooth avoidance arcs around challenging semantic obstacles like glass and carpets (common failure points during training), and maintains safe margins around corners (e.g., cabinet legs) without jitter. This high convergence indicates strong model confidence and stable decision-making. Conversely, the NavDP baseline (bottom row, red) exhibits severe planning failures. It fails to maintain safety margins, planning paths directly through glass walls or aggressively cutting corners into cabinet legs. Furthermore, its predicted trajectory bundles are highly divergent and disorganized, reflecting confused decision-making that struggles to converge on a safe path, often leading the robot to get stuck.

\subsection{Robustness and Zero-Shot Transfer}
We conduct two out-of-distribution (OOD) experiments to test generalization beyond the training distribution.

\textbf{1) OOD Geometric Generalization.} We utilize the 10 ``Hard'' scenes from the Scene-N1 dataset \cite{scene_n1}, which feature abstract geometric primitives (cylinders, cubes) placed at high density, with no residential furniture. On 1,010 tasks, our fine-tuned model achieves 90.1\% SR and 0.861 SPL, compared to 88.1\% SR and 0.851 SPL for the pretrained baseline. This suggests that the fine-tuned policy has learned obstacle avoidance skills that are not strictly tied to specific object categories.

\begin{figure}[h]
\centering
\includegraphics[width=\linewidth]{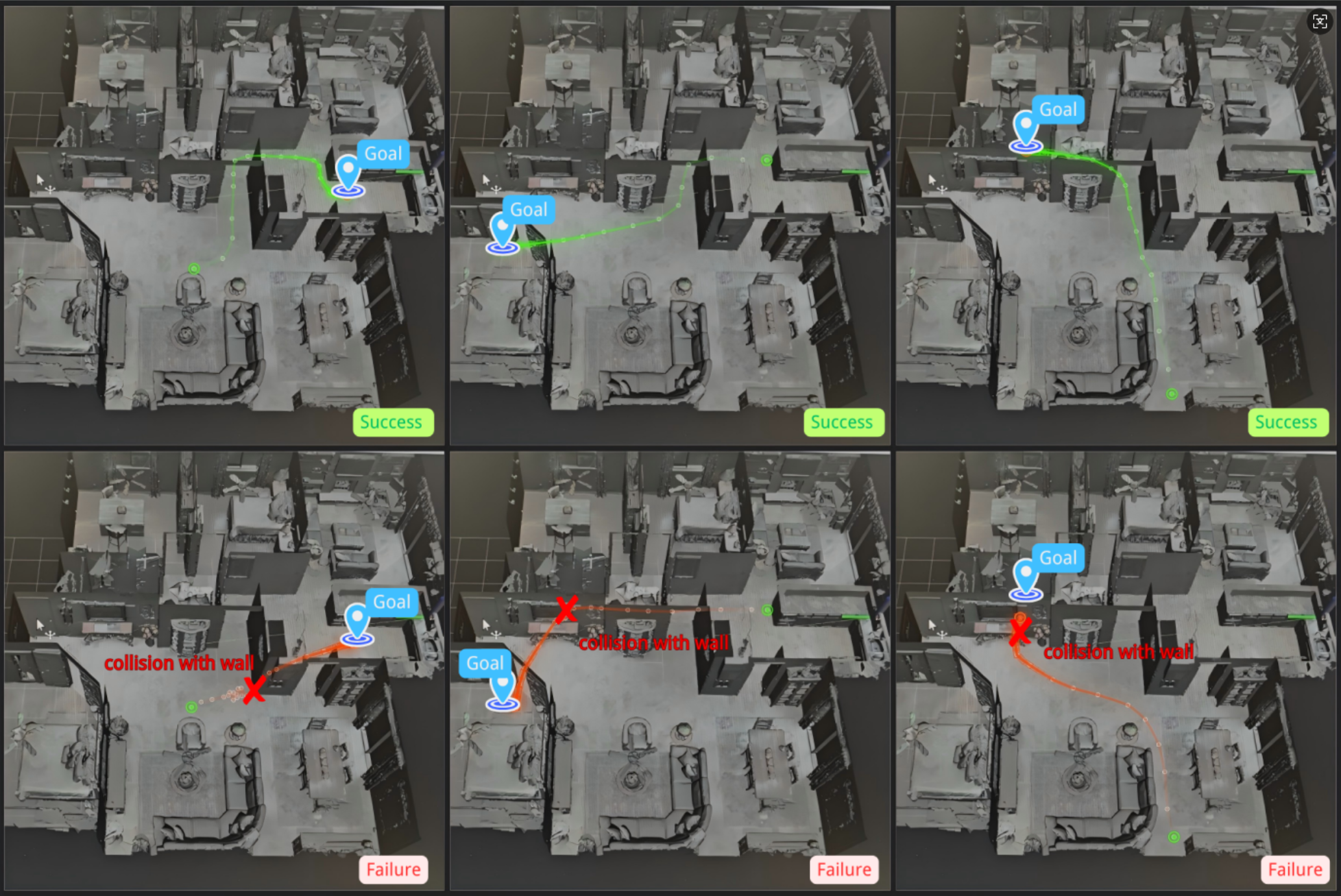}
\caption{Zero-shot sim-to-sim transfer in Habitat. Our fine-tuned policy (top row) successfully navigates across diverse scenarios, while the NavDP baseline (bottom row) consistently fails.}
\label{fig:habitat_transfer}
\end{figure}

\textbf{2) Zero-Shot Transfer to Habitat.} We deploy the Isaac Sim-trained policy directly in Habitat without any adaptation (Fig.~\ref{fig:habitat_transfer}). Zero-shot cross-simulator generalization is an emergent capability of the pretrained NavDP model; our RL fine-tuning does not enhance this capability per se, but reinforces safety-aware behaviors---maintaining wall clearance and avoiding near-misses---that demonstration data alone does not reliably instill. As a result, our fine-tuned policy (top row, green) consistently produces smooth, collision-free paths, while the NavDP baseline (bottom row, red) frequently exhibits path oscillation, aggressive corner-cutting, and direct wall collisions.

\subsection{Ablation Study}

\subsubsection{Effect of Fine-Tuning Strategy}

We evaluate the impact of different fine-tuning strategies on adaptation performance.

\begin{table}[h]
\centering
\caption{Effect of fine-tuning different model components.}
\resizebox{0.8\linewidth}{!}{
\begin{tabular}{ccc}
\hline
\textbf{Fine-Tuning Strategy} & \textbf{Success Rate} $\uparrow$ & \textbf{SPL} $\uparrow$ \\
\hline
\textbf{Ours (RL-FT, FT5)} & $\mathbf{0.587 \pm 0.014}$ & $\mathbf{0.537 \pm 0.002}$ \\
FT2 (2 layers) & $0.573 \pm 0.014$ & $0.534 \pm 0.017$ \\
FT-All (full model) & $0.520 \pm 0.066$ & $0.476 \pm 0.073$ \\
\hline
\end{tabular}
}
\label{tab:finetune_ablation}
\end{table}

As shown in Table~\ref{tab:finetune_ablation}, our selective strategy (FT5) yields the best performance. Fine-tuning too few parameters (FT2) bottlenecks the model's capacity to adapt. Conversely, updating the entire model (FT-All) yields no improvement over the pretrained NavDP baseline (0.520 vs.\ 0.520), and notably exhibits substantially higher variance ($\pm 0.066$ vs.\ $\pm 0.014$). This suggests that unconstrained full-model updates destabilize the RL training process: with all parameters free to change, the competing pressures of reward maximization and preserving pretraining representations lead to volatile optimization, erasing the gains that selective fine-tuning reliably achieves.

\subsubsection{Effect of Denoising Steps $K$ for Importance Ratio}

We investigate the effect of the number of denoising steps $K$ used to compute the importance ratio (Eq.~9). Rather than summing over all denoising steps, we only use the last $K$ steps of the denoising chain, as the final steps contribute most to the output trajectory. We vary $K \in \{3, 5, 7, 10\}$.

\begin{table}[h]
\centering
\caption{Effect of denoising steps $K$ on navigation performance.}
\resizebox{0.8\linewidth}{!}{
\begin{tabular}{ccc}
\hline
\textbf{Denoising Steps $K$} & \textbf{Success Rate} $\uparrow$ & \textbf{SPL} $\uparrow$ \\
\hline
$K=3$ & $0.567 \pm 0.020$ & $0.517 \pm 0.014$ \\
$K=5$ & $0.569 \pm 0.009$ & $0.520 \pm 0.004$ \\
$\mathbf{K=7}$ & $\mathbf{0.587 \pm 0.014}$ & $\mathbf{0.537 \pm 0.002}$ \\
$K=10$ & $0.532 \pm 0.015$ & $0.490 \pm 0.012$ \\
\hline
\end{tabular}
}
\label{tab:k_ablation}
\end{table}

Table~\ref{tab:k_ablation} shows that using the final $K=7$ steps is optimal. Using too few steps ($K=3, 5$) provides an incomplete estimation of the trajectory likelihood, weakening the policy gradient. Conversely, including all steps ($K=10$) degrades performance, as the earliest denoising steps are dominated by pure Gaussian noise, which introduces high variance into the importance ratio calculation without providing meaningful geometric feedback.

\subsubsection{Ablation of GRPO Components}

We further analyze the contribution of different components in the GRPO objective.

\begin{table}[h]
\centering
\caption{Ablation of GRPO objective components.}
\resizebox{0.8\linewidth}{!}{
\begin{tabular}{ccc}
\hline
\textbf{Method} & \textbf{Success Rate} $\uparrow$ & \textbf{SPL} $\uparrow$ \\
\hline
\textbf{Ours (RL-FT)} & $\mathbf{0.587 \pm 0.014}$ & $\mathbf{0.537 \pm 0.002}$ \\
Ours w/o Clipping & $0.553 \pm 0.014$ & $0.500 \pm 0.009$ \\
Ours w/o Adv. Norm. & $0.565 \pm 0.024$ & $0.516 \pm 0.018$ \\
\hline
\end{tabular}
}
\label{tab:objective_ablation}
\end{table}

As shown in Table~\ref{tab:objective_ablation}, both clipping and advantage normalization contribute significantly to performance. Removing either one leads to a noticeable drop, confirming their importance for stable training.

\section{Real-World Deployment}

To test our framework in physical settings, we deploy the fine-tuned policy on a quadruped robot \cite{omahoney2025offline}. We describe the system architecture, the sim-to-real transfer setup, and results.

\subsection{System Architecture and Asynchronous Deployment}
We use a Unitree Go2 quadruped robot with an NVIDIA Jetson AGX Orin and an Orbbec Gemini 336L camera. Deploying diffusion policies on edge devices is challenging due to the cost of iterative denoising. We address this with a ROS-based asynchronous framework that decouples planning from control.

The diffusion planner runs at about 2\,Hz on the Orin, outputting $H=24$ waypoints. A separate MPC tracker runs at 25\,Hz, with a prediction and control horizon of $N_{mpc}=15$ steps (1.5\,s), continuously following the latest trajectory and producing velocity commands. This ensures smooth motion even when planner latency varies.

\subsection{Sim-to-Real and Cross-Embodiment Transfer}
The policy was trained entirely in Isaac Sim on a wheeled platform (Clearpath Dingo) and is directly transferred to the Unitree Go2 quadruped without any real-world fine-tuning. This cross-embodiment transfer is enabled by the decoupled architecture: the diffusion policy outputs platform-agnostic waypoints, while a separate low-level MPC controller converts them into platform-specific velocity commands, confining embodiment-specific dynamics entirely to the controller.

\subsection{Experimental Setup and Testing Protocol}
We conduct experiments in two challenging indoor environments: a \textbf{Narrow Corridor} and a \textbf{Cluttered Office}. We compare the pretrained NavDP baseline against our RL fine-tuned model (Ours).
\begin{itemize}
    \item \textbf{Protocol:} Each method is evaluated over 20 independent trials (10 per scene).
    \item \textbf{Diverse Layouts:} To verify that the robot relies on real-time perception rather than map memorization, we randomly reconfigure obstacle positions (e.g., chairs, boxes) between every trial.
    \item \textbf{Failure Criteria:} A trial is marked as a failure immediately upon any collision.
\end{itemize}

\subsection{Real-World Navigation Performance}
Results are shown in Table~\ref{tab:realworld}. Our fine-tuned model achieves \textbf{70\%} success (14/20), compared to \textbf{50\%} (10/20) for the pretrained baseline.

\begin{figure}[h]
\centering
\includegraphics[width=\linewidth]{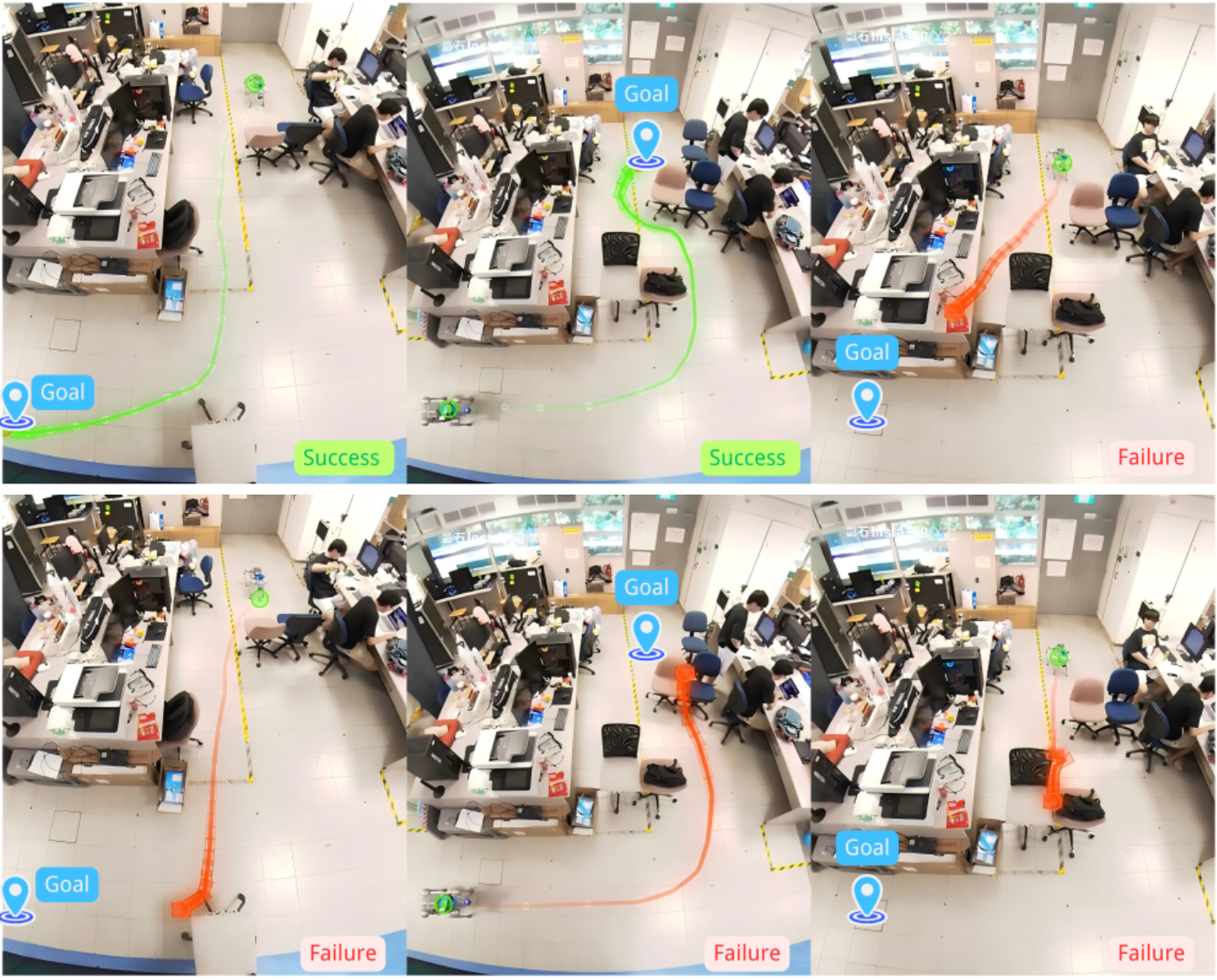}
\caption{Real-world navigation timelapse. Green and orange trajectories denote the actual executed paths of successful and failed runs, respectively. Our policy (top row) navigates smoothly, while the NavDP baseline (bottom row) struggles in dense clutter.}
\vspace{-10pt}
\label{fig:real_timelapse}
\end{figure}

\textbf{Qualitative Analysis.} As shown in Fig.~\ref{fig:real_timelapse}, real-world deployment on the quadruped robot reveals distinct behavioral differences. Our fine-tuned model (top row, green) demonstrates fluid obstacle avoidance, proactively steering clear of desk and chair legs (left). In dense obstacle clusters, it executes smooth, S-shaped avoidance maneuvers while consistently maintaining a safe clearance (middle). However, it occasionally fails in highly complex scenes, such as attempting to navigate under a table where clearance is insufficient (right), which we believe can be mitigated by further reinforcement learning feedback with reward terms that explicitly penalize low-clearance trajectories. Conversely, the pretrained baseline (bottom row, red) exhibits unstable and risky motion. It makes sharp, sudden turns near obstacles, leaving inadequate safety margins near corners (left). In dense areas, its trajectories jitter and bend irregularly, reflecting rapidly oscillating local decisions (middle). Notably, the baseline often attempts to traverse narrow gaps between chairs (right), exposing a critical lack of embodiment awareness—it mistakenly treats any visual gap as a traversable path without accounting for the robot's physical dimensions, leading to collisions.

\begin{table}[h]
\centering
\caption{Real-world navigation performance on Unitree Go2.}
\footnotesize
\resizebox{0.6\linewidth}{!}{
\begin{tabular}{cc}
\hline
\textbf{Method} & \textbf{Success Rate} $\uparrow$ \\
\hline
NavDP (Pretrained) & 50\% (10/20) \\
\textbf{Ours (RL-FT)} & \textbf{70\% (14/20)} \\
\hline
\end{tabular}
}
\label{tab:realworld}
\end{table}


\section{Conclusion}

This paper presents a reinforcement learning fine-tuning framework to adapt pretrained diffusion-based navigation policies without requiring additional expert demonstrations. By formulating generation as a trajectory-level MDP and leveraging Group Relative Policy Optimization (GRPO), our method achieves memory-efficient, critic-free adaptation. More broadly, our work demonstrates that RL fine-tuning serves as an effective pathway to bridge the gap between learned navigation models and safe real-world deployment: by introducing collision penalties and clearance constraints through online interaction, the framework instills safety-aware planning behaviors that offline imitation learning often struggles to ensure in novel domains. Furthermore, we demonstrate that selectively freezing the perception backbone and lower Transformer layers is critical to preventing catastrophic forgetting of zero-shot visual priors while enabling rapid geometric adaptation. Extensive experiments show that our framework significantly improves both success rates and navigation safety in unseen environments, producing more reliable, collision-free trajectories, and seamlessly transfers to physical quadruped robots without real-world training.

While our approach robustly enhances local obstacle avoidance and planning safety, experiments also reveal inherent limitations in current end-to-end reactive paradigms: they lack multi-step global routing for long-horizon navigation and occasionally struggle with fine-grained embodiment awareness in tight spaces. Addressing these bottlenecks through integrated topological planning and dynamic entity-aware rewards remains a crucial direction for scalable, safe real-world deployment.

\addtolength{\textheight}{-12cm}
\bibliographystyle{IEEEtran}
\bibliography{citation} 

\end{document}